\documentclass[11pt,a4paper]{article}
\pdfoutput=1
\usepackage{naaclhlt2018}
\usepackage{times}
\usepackage{latexsym}
\usepackage{microtype}
\usepackage{amsmath}
\usepackage{amsfonts}
\usepackage{mathtools}
\usepackage{booktabs}
\usepackage{url}
\usepackage{adjustbox}
\usepackage{xcolor}

\definecolor{awesome}{rgb}{1.0, 0.13, 0.32}
\definecolor{blue(pigment)}{rgb}{0.2, 0.2, 0.6}
\definecolor{darkcyan}{rgb}{0.0, 0.55, 0.55}
\definecolor{electricviolet}{rgb}{0.56, 0.0, 1.0}

\DeclareFontFamily{U}{rcjhbltx}{}
\DeclareFontShape{U}{rcjhbltx}{m}{n}{<->rcjhbltx}{}
\DeclareSymbolFont{hebrewletters}{U}{rcjhbltx}{m}{n}

\DeclareMathSymbol{\lamed}{\mathord}{hebrewletters}{108}

\usepackage{pgfplots}

\usepackage{cleveref}
\crefname{section}{\S}{\S\S}
\Crefname{section}{\S}{\S\S}
\crefname{table}{Table}{}
\crefname{figure}{Fig.}{}
\crefname{algorithm}{Alg.}{}
\crefname{equation}{eq.}{}
\crefname{appendix}{Appendix}{}
\crefformat{section}{\S#2#1#3}

\newcommand{\vtheta}{{\boldsymbol \theta}}
\newcommand{\vv}{\mathbf{v}}
\newcommand{\vu}{\mathbf{u}}
\newcommand{\ptheta}{p_\vtheta}

\newcommand{\saveforCR}[1]{}

\usepackage[T1]{fontenc}
\usepackage{la}  
\usepackage{courier}  
\newcommand{\defn}[1]{\textbf{#1}}
\newcommand{\lexm}[1]{\text{\textla{\Large #1}}}
\newcommand{\word}[1]{\textsf{\small #1}}
\newcommand{\pos}[1]{\textsc{#1}}
\newcommand{\slot}[1]{\textsc{#1}}
\newcommand{\attval}[2]{\mbox{\slot{#1}$=$\slot{#2}}}
\newcommand{\angles}[1]{\langle #1 \rangle}
\newcommand{\brackets}[1]{\big[ #1 \big]}

\aclfinalcopy 
\def\aclpaperid{993} 

\title{Unsupervised Disambiguation of Syncretism in Inflected Lexicons}
\author{Ryan Cotterell {\;\normalfont and\;} Christo Kirov {\;\normalfont and\;} Sabrina J. Mielke {\;\normalfont and\;} Jason Eisner\\
  Department of Computer Science, Johns Hopkins University\\
  {\tt \{ryan.cotterell@,ckirov1@,sjmielke@,jason@cs\}.jhu.edu} \\}

\date{}

\begin{document}

\thispagestyle{plain}
\pagestyle{plain}

\maketitle
\begin{abstract}
  Lexical ambiguity makes it difficult to compute various useful statistics of
  a corpus.  A given word form might represent any of several
  morphological feature bundles.  One can, however, use unsupervised
  learning (as in EM) to fit a model that probabilistically
  disambiguates word forms.  We present such an approach, which
  employs a neural network to smoothly model a prior distribution over
  feature bundles (even rare ones).  Although this basic model does
  not consider a token's context, that very property allows it to
  operate on a simple list of unigram type counts, partitioning each
  count among different analyses of that unigram.
  We discuss evaluation metrics for this novel
  task and report results on 5 languages.
\end{abstract}

\section{Introduction}\label{sec:introduction}
Inflected lexicons---lists of morphologically inflected forms---are commonplace in NLP.
Such lexicons currently exist for over 100 languages in a standardized annotation scheme
\cite{KIROV18}, making them one of the most multi-lingual annotated
resources in existence. These lexicons are typically annotated at the
type level, i.e., each word type is listed with its \emph{possible}
morphological analyses, divorced from sentential context.

One might imagine that most word types are unambiguous.  However, many
inflectional systems are replete with a form of ambiguity termed
syncretism---a systematic merger of morphological slots.  In English,
some verbs have five distinct inflected forms, but regular verbs (the
vast majority) merge two of these and so distinguish only
four. The verb \lexm{sing} has the past tense form
\word{sang} but the participial form \word{sung}; the verb
\lexm{talk}, on the other hand, employs \word{talked} for both
functions.  The form \word{talked} is, thus, said to be syncretic. Our task is to
partition the count of
\word{talked} in a corpus between the past-tense and participial readings,
respectively.

In this paper, we model a generative probability distribution over
\emph{annotated} word forms, and fit the model parameters using the
token counts of \emph{unannotated} word forms.  The resulting
distribution predicts how to partition each form's token count among
its possible annotations.  While our method actually deals with all ambiguous forms
in the lexicon, it is particularly useful for syncretic forms because
syncretism is often systematic and pervasive.

In English, our unsupervised procedure learns from the counts of
irregular pairs like \word{sang}--\word{sung} that a verb's past tense
tends to be more frequent than its past participle.  These learned
parameters are then used to disambiguate \word{talked}.  The method
can also learn from regular paradigms.  For example, it learns from
the counts of pairs like \word{runs}--\word{run} that singular
third-person forms are common.  It then uses these learned parameters
to guess that tokens of \word{run} are often singular or third-person
(though never both at once, because the lexicon does not
list that as a possible analysis of \word{run}).

\section{Formalizing Inflectional Morphology}

\begin{table}
\centering
\begin{tabular}{lllll}
  \toprule
  & {\sc sg} & {\sc pl} & {\sc sg} & {\sc pl}  \\[.2em] \midrule
      {\sc nom} & \textbf{\color{electricviolet}{\word{Wort}}} & \textbf{\color{darkcyan}{\word{W{\"o}rter}}} & \word{Herr} & \textbf{\color{awesome}{\word{Herren}}} \\[.2em]
    {\sc gen} & \word{Wortes} & \textbf{\color{darkcyan}{\word{W{\"o}rter}}} & \textbf{\color{blue(pigment)}{\word{Herrn}}} & \textbf{\color{awesome}{\word{Herren}}} \\[.2em]
{\sc acc} &  \textbf{\color{electricviolet}{\word{Wort}}} & \textbf{\color{darkcyan}{\word{W{\"o}rter}}}  & \textbf{\color{blue(pigment)}{\word{Herrn}}} & \textbf{\color{awesome}{\word{Herren}}} \\[.2em]
{\sc dat}  & \word{Worte} & \word{W{\"o}rtern} & \textbf{\color{blue(pigment)}{\word{Herrn}}} & \textbf{\color{awesome}{\word{Herren}}} \\[.2em]
\bottomrule
\end{tabular}
\caption{Full paradigms for the German nouns \lexm{Wort} (``word'') and \lexm{Herr} (``gentleman'') with abbreviated and tabularized UniMorph
annotation. The syncretic forms
are bolded and colored by ambiguity class. Note that, while in the plural the nominative and accusative are always syncretic across all paradigms, the same
is not true in the singular.
}
\label{tab-paradigm}
\end{table}
We adopt the framework of word-based morphology
\cite{aronoff1976word,spencer1991morphological}.
In the present paper, we consider only inflectional morphology.  An
\defn{inflected lexicon} is a set of word types. Each \defn{word type}
is a 4-tuple of a part-of-speech tag, a lexeme, an inflectional slot,
and a surface form.

A \defn{lexeme} is a discrete object (represented by an arbitrary integer or string, which we typeset in \lexm{cursive}) that indexes the word's core meaning and part of speech.  A \defn{part-of-speech (POS) tag} is a coarse syntactic category such as \pos{Verb}.  Each POS tag allows some set of lexemes, and also allows some set of inflectional \defn{slots} such as ``1st-person present singular.''  Each allowed $\angles{\text{tag, lexeme, slot}}$ triple is realized---in only one way---as an inflected \defn{surface form}, a string over a fixed phonological or orthographic alphabet $\Sigma$.  In this work, we take $\Sigma$ to be an orthographic alphabet.

A \defn{paradigm} $\pi(t, \ell)$ is the mapping from tag $t$'s slots to the surface forms that ``fill'' those slots for lexeme $\ell$.  For example, in the English paradigm $\pi(\pos{Verb}, \lexm{talk})$, the past-tense slot is said to be filled by \word{talked}, meaning that the lexicon contains the tuple $\angles{\pos{Verb}, \lexm{talk}, \slot{past}, \word{talked}}$.\footnote{Lexicographers will often refer to a paradigm by its \defn{lemma}, which is the surface form that fills a certain designated slot such as the infinitive.  We instead use lexemes because lemmas may be ambiguous: \scalebox{.85}{\word{bank}} is the lemma for at least two nominal and two verbal paradigms.}

We will specifically work with the UniMorph annotation scheme
\cite{sylak2016composition}.  Here each slot specifies a
morpho-syntactic bundle of inflectional features (also called
a morphological tag in the literature), such as tense, mood, person, number,
and gender.  For example, the German surface form \word{W{\"o}rtern}
is listed in the lexicon with tag \pos{Noun}, lemma \lexm{Wort}, and a slot specifying the
feature bundle $\brackets{\attval{num}{pl}, \attval{case}{dat}}$. An example of
UniMorph annotation is found in \cref{tab-paradigm}.

\subsection{What is Syncretism?}\label{sec:ambiguity1}

We say that a surface form $f$ is \defn{syncretic} if two slots $s_1 \neq s_2$ exist
such that some paradigm $\pi(t,\ell)$ maps both $s_1$ and $s_2$ to $f$.
In other words, a single form fills
multiple slots in a paradigm: syncretism may
be thought of as intra-paradigmatic ambiguity. This definition
does depend on the exact annotation scheme in use, as some schemes collapse
syncretic slots.  For example, in German nouns,
{\em no} lexeme distinguishes the nominative, accusative and genitive plurals.
Thus, a human-created lexicon might employ a single slot
$\brackets{\attval{num}{pl}, \attval{case}{nom/acc/gen}}$ and say that
\word{W{\"o}rter} fills just this slot rather than three separate slots.
For a discussion, see \newcite{baerman2005syntax}.

\subsection{Inter-Paradigmatic Ambiguity}\label{sec:ambiguity2}
A different kind of ambiguity occurs
when a surface form belongs to more
than one paradigm. A form
$f$ is inter-paradigmatically ambiguous if
$\angles{ t_1, \ell_1, s_1, f }$ and $\angles{ t_2, \ell_2, s_2, f}$ are both in the lexicon for lexemes $\angles{t_1,\ell_1} \neq \angles{t_2,\ell_2}$.

For example, \word{talks}
belongs to the English paradigms $\pi(\pos{Verb},\lexm{talk})$ and $\pi(\pos{Noun}, \lexm{talk})$. The model we present in \cref{sec:model} will resolve both syncretism and inter-paradigmatic ambiguity. However, our exposition focuses on the former, as it is cross-linguistically more common.

\subsection{Disambiguating Surface Form Counts}
The previous  sections \cref{sec:ambiguity1} and
\cref{sec:ambiguity2} discussed two types of ambiguity found in inflected lexicons.
The goal of this paper is the \emph{disambiguation}
of raw surface form counts, taken
from an unannotated text corpus. In other words, given
such counts, we seek to impute
the fractional counts for individual lexical entries (4-tuples), which are unannotated in raw text.
Let us assume that the  word
\word{talked} is observed $c\left(\text{\word{talked}}\right)$ times in a
raw English text corpus.
We do not
know which instances of \word{talked} are participles and which are
past tense forms. However, given a probability distribution
$\ptheta(t, \ell, s \mid f)$, we may disambiguate these counts in
expectation, i.e., we attribute a count of
\begin{equation*}\label{eq:e-step}
c\left(\text{\word{talked}}\right)
\cdot \ptheta( \pos{Verb}, \lexm{talk}, \slot{past\_part} \mid \text{\word{talked}})
\end{equation*}
to the past participle of the \pos{Verb} \lexm{talk}. Our aim is the construction and unsupervised
estimation of the distribution $\ptheta(t, \ell, s \mid f)$.

While the task at hand is novel, what applications does it have?  We are especially interested in \emph{sampling} tuples $\angles{
t, \ell, s, f }$ from an inflected lexicon.
Sampling is a necessity for creating train-test splits for evaluating
morphological inflectors, which has recently become a standard task
in the literature \cite{DurrettDeNero2013,hulden-forsberg-ahlberg:2014:EACL,nicolai-cherry-kondrak:2015:NAACL-HLT,faruqui:2016:infl},
and has seen two shared tasks
\cite{cotterell-EtAl:2016:SIGMORPHON,cotterell-conll-sigmorphon2017}.
Creating train-test splits for training inflectors
involves sampling \emph{without replacement} so that all
test types are unseen.
Ideally, we would like more frequent word types in the training portion and less frequent ones in the test portion. This is a realistic evaluation: a training lexicon for a new language would tend to contain frequent types, so the system should be tested on its
ability to extrapolate to rarer types that could not be looked up in
that lexicon, as discussed by \newcite{cotterell-peng-eisner-2015}.
To make the split, we sample $N$ word types without replacement, which is equivalent to collecting the first $N$ distinct forms from an annotated corpus generated from the same unigram distribution.

The fractional counts that our method estimates may also be useful for
corpus linguistics---for example, tracking the frequency of specific lexemes
over time, or comparing the rate of participles in the work of two different authors.

Finally, the fractional counts can aid the training of NLP methods that operate on a raw corpus, such as distributional embedding of surface form types into a vector space.  Such methods sometimes consider the morphological properties (tags, lexemes, and slots) of nearby context words.  When the morphological properties of a context word $f$ are ambiguous, instead of tagging (which may not be feasible) one could {\em fractionally} count the occcurrences of the possible analyses according to $\ptheta(t, \ell, s \mid f)$, or else characterize $f$'s morphology with a single {\em soft} indicator vector whose elements are the probabilities of the properties according to $\ptheta(t, \ell, s \mid f)$.

\section{A Neural Latent Variable Model}\label{sec:model}
In general, we will only observe unannotated word forms $f$.
We model these as draws from a distribution over form types $\ptheta(f)$, which marginalizes out
the unobserved structure of the lexicon---which tag, lexeme and slot generated each
form. Training the parameters of this latent-variable model will recover the posterior distribution over analyses of a form, $\ptheta(t, \ell, s \mid f)$,
which allows us to disambiguate counts at the type level.

The latent-variable model is a Bayesian network,$\!\!$
\begin{equation}\label{eq:latent-variable-model}
\ptheta(f) = \sum_{\mathclap{\angles{ t, \ell, s } \in {\cal T} \times {\cal L} \times {\cal S}}}  \ptheta(t)\, \ptheta(\ell \mid t)\, \ptheta(s \mid t)\, \delta(f \mid t,\ell,s)
\end{equation}
where ${\cal T}, {\cal L}, {\cal S}$ range over the possible tags, lexemes, and slots of the language, and $\delta(f \mid t,\ell,s)$
returns 1 or 0 according to whether the lexicon lists $f$ as the (unique) realization of $\angles{ t, \ell, s }$.  We fix $p_\theta(s \mid t) = 0$ if the lexicon lists no tuples of the form $\angles{t,\cdot,s,\cdot}$, and otherwise model
\begin{equation}
  \ptheta(s \mid t) \propto \exp\left(\vu ^{\top} \tanh\left(\mathbf{W} \cdot \vv_{t,s} \right) \right) > 0 \label{eq:nn}
\end{equation}
where $\vv_{t,s}$ is a multi-hot vector whose ``1'' components indicate the morphological features possessed by $\angles{t,s}$: namely attribute-value pairs such as $\attval{pos}{Verb}$ and $\attval{num}{pl}$.  Here $\vu \in \mathbb{R}^d$ and $\mathbf{W}$ is a conformable matrix of weights.  This formula specifies a neural network with $d$ hidden units, which
can learn to favor or disfavor specific soft conjunctions of morphological features.
Finally, we define $\ptheta(t) \propto \exp \omega_t$ for $t \in {\cal T}$, and $\ptheta(\ell \mid t) \propto \exp \omega_{t,\ell}$ or $0$ if the lexicon lists no tuples of the form $\angles{t,\ell,\cdot,\cdot}$.  The model's parameter vector $\vtheta$ specifies $\vu, \mathbf{W}$, and the $\omega$ values.\looseness=-1

\subsection{Inference and Learning}
We maximize the regularized log-likelihood
\begin{equation}\label{eq:loglik}
\sum_{f \in {\cal F}} c(f) \log \ptheta(f) + \frac{\lambda}{2} ||\vtheta||_2^2
\end{equation}
where ${\cal F}$ is the set of surface form types and $p_\theta(f)$ is defined by \eqref{eq:latent-variable-model}.
It is straightforward to use a gradient-based optimizer, and we do.
However, \eqref{eq:loglik} could also be maximized by an intuitive EM algorithm: at each iteration, the E-step uses the current model parameters to partition each count $c(f)$ among possible analyses, as in \eqref{eq:e-step}, and then the M step improves the parameters by following the gradient of {\em supervised} regularized log-likelihood as if it had observed those fractional counts.

On each iteration, either algorithm loops through all listed $(t,s)$
pairs, all listed $(t,\ell)$ pairs, and all observed forms $f$, taking
time at most proportional to the size of the lexicon.  In practice,
training completes within a few minutes on a modern
laptop.

\subsection{Baseline Models}

To the best of our knowledge, this
disambiguation task is novel.
Thus, we resort to comparing three variants of our
model in lieu of a previously published baseline.
We evaluate three simplifications of the slot model, to
investigate whether the complexity of equation~\eqref{eq:nn} is
justified.
\begin{description}
  \item[\sc unif:] $p(s \mid t)$ is uniform over the slots $s$ that are
  listed with $t$.  This involves no learning.
  \item[\sc free:] $p(s \mid t) \propto \exp \omega_{t,s}$: a model with a single parameter $\omega_{t,s} \in \mathbb{R}$ per slot.  This can capture any distribution, but it has less inductive bias: slots that share morphological features do not share parameters.
  \item[\sc linear:] $p(s \mid t) \propto \exp (\vu^{\top} \vv_{t,s})$: a linear model with no
  conjunctions between morphological features.  This chooses the features
  orthogonally, in the sense that (e.g.) if verbal paradigms have a
  complete 3-dimensional grid of slots indexed by their \slot{person}, \slot{num}, and \slot{tense} attributes, then sampling from $p(s \mid \pos{Verb})$ is equivalent to independently sampling these three coordinates.  Moreover, $p(\attval{num}{pl} \mid \pos{Noun}) = p(\attval{num}{pl} \mid \pos{Verb})$.
\end{description}

\section{Experiments}

\subsection{Computing Evaluation Metrics}

We first evaluate \defn{perplexity}. Since our model is a tractable generative model, we may easily evaluate its perplexity on held-out tokens.
 For each language, we randomly partition the observed surface tokens into 80\% training, 10\% development, and 10\% test.  We then estimate the parameters of our model
by maximizing \eqref{eq:loglik} on the counts from the training portion, selecting hyperparameters such that the estimated parameters%
\footnote{Our vocabulary and parameter set are determined from the {\em lexicon}.  Thus we create a regularized parameter $\omega_\ell$, yielding a smoothed estimate $p(\ell)$, even if the training count $c(\ell) = 0$.}
minimize perplexity on the development portion.  We then report perplexity on the test portion.

Using the same hyperparameters, we now train our latent-variable model $\ptheta$
{\em without} supervision on 100\% of the observed surface forms
$f$. We now measure how poorly, for the average surface form type $f$, we recovered the
maximum-likelihood distribution \mbox{$\hat{p}(t,\ell,s \mid f)$}
that would be estimated {\em with} supervision in terms of \textbf{KL-divergence}:
\begin{align}
 \lefteqn{\sum_f \hat{p}(f)\; \text{KL}(\hat{p}(\cdot \mid f) \mid\mid \ptheta(\cdot \mid f))} \label{eq:KL}\\
 &= \frac{1}{N}\sum_{i=1}^N \log_2 \frac{\hat{p}(t_i,\ell_i,s_i \mid f_i)}{\ptheta(t_i,\ell_i,s_i \mid f_i)} \nonumber
\end{align}
We can see that this formula reduces to a simple average over disambiguated tokens $i$.

\subsection{Training Details and Hyperparameters}

We optimized on training data using batch gradient descent with a fixed learning rate.  We used perplexity on development data to jointly choose the learning rate, the initial random $\vtheta$ (from among several random restarts), the regularization coefficient $\lambda \in \{10^{-1}, 10^{-2}, 10^{-3}, 10^{-4} \}$ and the neural network architecture.
The {\sc neural} architecture shown in \cref{eq:nn} has 1 hidden layer,
but we actually generalized this to consider networks with $k \in \{1, 2, 3, 4\}$ hidden layers of $d=100$ units each.  In some cases, the model selected on development data had $k$ as high as 3.  Note that the {\sc linear} model corresponds to $k=0$.

\subsection{Datasets}

Each language constitutes a separate experiment.  In each case we
obtain our lexicon from the UniMorph project and our surface form
counts from Wikipedia.  To approximate supervised counts to estimate
$\hat{p}$ in the KL evaluation, we analyzed the surface form tokens in
Wikipedia (in context) using the tool in \newcite{straka2016udpipe},
as trained on the disambiguated Universal Dependencies (UD) corpora
\cite{NIVRE16.348}.  We wrote a script\footnote{The script discarded up to 31\% of the tokens
  because the UD analysis could not be successfully converted into an
  UniMorph analysis that was present in the lexicon.} to convert the resulting
analyses from UD format into $\angles{t,\ell,s,f}$ tuples in UniMorph
format for five languages---Czech (cs), German (de),
Finnish (fi), Hebrew (he), Swedish (sv)---each of which displays
both kinds of ambiguity in its UniMorph lexicon.
Lexicons with these approximate supervised counts are provided as
supplementary material.

\subsection{Results}
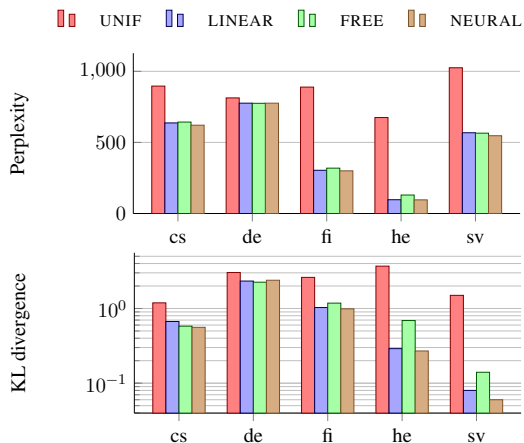
\begin{figure}
  \centering
  \scalebox{.7}{
    \begin{tikzpicture}
      \begin{axis}[
          axis x line*=bottom, axis y line*=left,
          ybar=0pt,
          ymin=0,
          ylabel={Perplexity},
          y label style={at={(axis description cs:-0.125,0.5)}},
          bar width=7pt,
          ymajorgrids,
          yminorgrids,
          symbolic x coords={cs,de,fi,he,sv},
          enlarge x limits={value=0.15, auto},
          xtick=data,
          xticklabel style={text height=.7em},
          nodes near coords align={horizontal}, every node near coord/.append style={rotate=90},
          width=23em, height=12em,
          legend style={at={(0.41,1.1)},anchor=south,draw=none,fill=none,column sep=1em,name=legend},
          legend columns=4,
          legend cell align=left
          ]
        \addplot[  red!50!black,fill=  red!50!white] coordinates {(cs,896) (de,813) (fi,889) (he,675) (sv,1025)};
        \addplot[ blue!35!black,fill= blue!35!white] coordinates {(cs,637) (de,776) (fi,304) (he,97) (sv,568)};
        \addplot[green!35!black,fill=green!35!white] coordinates {(cs,643) (de,775) (fi,319) (he,130) (sv,565)};
        \addplot[brown!65!black,fill=brown!65!white] coordinates {(cs,621) (de,776) (fi,300) (he,96) (sv,547)};
        \legend{\textsc{unif},\textsc{linear},\textsc{free},\textsc{neural}}
      \end{axis}
    \end{tikzpicture}
  }
  \scalebox{.7}{
    \begin{tikzpicture}
      \begin{axis}[
          axis x line*=bottom, axis y line*=left,
          ybar=0pt,
          ymode=log,
          log origin=infty,
          ylabel={KL divergence},
          y label style={at={(axis description cs:-0.125,0.5)}},
          bar width=7pt,
          ymajorgrids,
          yminorgrids,
          symbolic x coords={cs,de,fi,he,sv},
          enlarge x limits={value=0.15, auto},
          xtick=data,
          xticklabel style={text height=.7em},
          nodes near coords align={horizontal}, every node near coord/.append style={rotate=90},
          width=23em, height=12em
          ]
        \addplot[  red!50!black,fill=  red!50!white] coordinates {(cs,1.19) (de,3.03) (fi,2.61) (he,3.69) (sv,1.5)};
        \addplot[ blue!35!black,fill= blue!35!white] coordinates {(cs,0.67) (de,2.33) (fi,1.03) (he,0.29) (sv,0.08)};
        \addplot[green!35!black,fill=green!35!white] coordinates {(cs,0.58) (de,2.25) (fi,1.18) (he,0.69) (sv,0.14)};
        \addplot[brown!65!black,fill=brown!65!white] coordinates {(cs,0.56) (de,2.39) (fi,0.99) (he,0.27) (sv,0.06)};
      \end{axis}
    \end{tikzpicture}
    \hspace*{.2em}
  }
  \caption{Unsupervised and supervised test results under each model, averaged over 50 training-dev-test splits.}
  \label{fig:results}
  \end{figure}

Our results are graphed in \cref{fig:results}, exact numbers are found in \cref{tab:results}. We find that the {\sc neural} model slightly outperforms the other baselines on languages except for
German.  The {\sc linear} model is quite competitive as well.

  \begin{table}[h]
    \centering
    \begin{adjustbox}{width=1\columnwidth}
      \begin{tabular}{l ll ll ll ll } \toprule
              & \multicolumn{2}{c}{\sc neural net}  & \multicolumn{2}{c}{\sc free} & \multicolumn{2}{c}{\sc linear}  & \multicolumn{2}{c}{\sc uniform} \\
        lang & perp & KL & perp & KL &  perp & KL &  perp & KL   \\ \cmidrule(lr){0-0} \cmidrule(lr){2-3} \cmidrule(lr){4-5} \cmidrule(lr){6-7} \cmidrule(lr){8-9}
        cs & \textbf{621}  & \textbf{0.56} & 643 & 0.58 & 637 & 0.67 & 896 & 1.19   \\
        de & 776  & 2.39 & \textbf{775} & \textbf{2.25} & 776 & 2.33 & 813 & 3.03   \\
        fi & \textbf{300}  & \textbf{0.99} & 319 & 1.18 & 304 & 1.03 & 889 & 2.61   \\
        he & \textbf{96}   & \textbf{0.27} & 130 & 0.69 & 97  & 0.29 & 675 & 3.69   \\
        sv & \textbf{547}  & \textbf{0.06} & 565 & 0.14 & 568 & 0.08 & 1025 & 1.5 \\
        \bottomrule
      \end{tabular}
    \end{adjustbox}
    \caption{Results for the best performing neural network (hyperparameters
      selected on dev) and the three baselines under both performance metrics. Best are bolded.}
    \label{tab:results}
  \end{table}

{\sc unif} would have a KL divergence of 0 bits if all forms were either unambiguous or uniformly ambiguous.  Its higher value means the unsupervised task is nontrivial.  Our other models substantially outperform {\sc unif}.  {\sc neural} matches the supervised distributions reasonably closely, achieving an average KL of $< 1$ bit on all languages but German.

\section{Related Work}
By far the closest work to ours is the seminal paper of
\newcite{baayen1996estimating}, who asked the following question:
``Given a form that is previously unseen in a sufficiently large
training corpus, and that is morphologically $n$-ways ambiguous [...]
what is the best estimator for the lexical prior probabilities for the
various functions of the form?''  While we address the same task,
i.e., estimation of a lexical prior, \newcite{baayen1996estimating}
assume supervision in the form of an disambiguated corpus. We are the
first to treat the specific task in an unsupervised fashion.
We discuss other work below.

\paragraph{Supervised Morphological Tagging.}
Morphological tagging is a common task in NLP; the state
of the art is currently held by neural models \cite{heigold-neumann-vangenabith:2017:EACLlong}.
This task is distinct
from the problem at hand. Even if a tagger obtains the possible
analyses from a lexicon, it is still trained in a supervised manner
to choose among analyses.

\paragraph{Unsupervised POS Tagging.}
Another vein of work that is similar to ours is that of unsupervised
part-of-speech (POS) tagging. Here, the goal is map sequences of forms
into coarse-grained syntactic categories.
\newcite{christodoulopoulos-goldwater-steedman:2010:EMNLP} provide a
useful overview of previous work. This task differs from ours on two
counts. First, we are interested in finer-grained morphological
distinctions: the universal POS tagset
\cite{PETROV12.274} makes 12 distinctions, whereas UniMorph has
languages expressing hundreds of distinctions. Second, POS tagging
deals with the induction of syntactic categories from sentential
context.

We note that purely unsupervised morphological tagging, has yet to be attempted to the best of our knowledge.

\section{Conclusion}
We have presented a novel generative latent-variable model for
resolving ambiguity in unigram counts, notably due to syncretism.  Given
a lexicon, an unsupervised model partitions the corpus count for each
ambiguous form among its analyses listed in a lexicon. We empirically
evaluated our method on 5 languages under two evaluation
metrics. The code is availabile at \url{https://sjmielke.com/papers/syncretism},
along with type-disambiguated unigram counts for all lexicons
provided by the UniMorph project (100+ languages).

\bibliography{morphimput}

\begin{thebibliography}{18}
\expandafter\ifx\csname natexlab\endcsname\relax\def\natexlab#1{#1}\fi

\bibitem[{Aronoff(1976)}]{aronoff1976word}
Mark Aronoff. 1976.
\newblock \emph{Word Formation in Generative Grammar}.
\newblock Number~1 in Linguistic Inquiry Monographs. MIT Press, Cambridge, MA.

\bibitem[{Baayen and Sproat(1996)}]{baayen1996estimating}
Harald Baayen and Richard Sproat. 1996.
\newblock Estimating lexical priors for low-frequency morphologically ambiguous
  forms.
\newblock \emph{Computational Linguistics}, 22(2):155--166.

\bibitem[{Baerman et~al.(2005)Baerman, Brown, and Corbett}]{baerman2005syntax}
Matthew Baerman, Dunstan Brown, and Greville~G. Corbett. 2005.
\newblock \emph{The Syntax-Morphology Interface: {A} study of Syncretism},
  volume 109.
\newblock Cambridge University Press.

\bibitem[{Christodoulopoulos et~al.(2010)Christodoulopoulos, Goldwater, and
  Steedman}]{christodoulopoulos-goldwater-steedman:2010:EMNLP}
Christos Christodoulopoulos, Sharon Goldwater, and Mark Steedman. 2010.
\newblock \href {http://www.aclweb.org/anthology/D10-1056} {Two decades of
  unsupervised {POS} induction: {H}ow far have we come?}
\newblock In \emph{Proceedings of the Conference on Empirical Methods in
  Natural Language Processing (EMNLP)}, pages 575--584, Cambridge, MA.
  Association for Computational Linguistics.

\bibitem[{Cotterell et~al.(2017)Cotterell, Kirov, Sylak-Glassman, Walther,
  Vylomova, Xia, Faruqui, K{\"u}bler, Yarowsky, Eisner, and
  Hulden}]{cotterell-conll-sigmorphon2017}
Ryan Cotterell, Christo Kirov, John Sylak-Glassman, G{\'e}raldine Walther,
  Ekaterina Vylomova, Patrick Xia, Manaal Faruqui, Sandra K{\"u}bler, David
  Yarowsky, Jason Eisner, and Mans Hulden. 2017.
\newblock \href {http://aclweb.org/anthology/K/K17/K17-2001.pdf} {The
  {CoNLL-SIGMORPHON} 2017 shared task: {U}niversal morphological reinflection
  in 52 languages}.
\newblock In \emph{Proceedings of the CoNLL-SIGMORPHON 2017 Shared Task:
  Universal Morphological Reinflection}, Vancouver, Canada. Association for
  Computational Linguistics.

\bibitem[{Cotterell et~al.(2016)Cotterell, Kirov, Sylak-Glassman, Yarowsky,
  Eisner, and Hulden}]{cotterell-EtAl:2016:SIGMORPHON}
Ryan Cotterell, Christo Kirov, John Sylak-Glassman, David Yarowsky, Jason
  Eisner, and Mans Hulden. 2016.
\newblock \href {http://anthology.aclweb.org/W16-2002} {The {SIGMORPHON} 2016
  shared task---morphological reinflection}.
\newblock In \emph{Proceedings of the 14th SIGMORPHON Workshop on Computational
  Research in Phonetics, Phonology, and Morphology}, pages 10--22, Berlin,
  Germany. Association for Computational Linguistics.

\bibitem[{Cotterell et~al.(2015)Cotterell, Peng, and
  Eisner}]{cotterell-peng-eisner-2015}
Ryan Cotterell, Nanyun Peng, and Jason Eisner. 2015.
\newblock \href {http://cs.jhu.edu/~jason/papers/#cotterell-peng-eisner-2015}
  {Modeling word forms using latent underlying morphs and phonology}.
\newblock \emph{Transactions of the Association for Computational Linguistics},
  3:433--447.

\bibitem[{Durrett and DeNero(2013)}]{DurrettDeNero2013}
Greg Durrett and John DeNero. 2013.
\newblock \href {http://www.aclweb.org/anthology/N13-1138} {Supervised learning
  of complete morphological paradigms}.
\newblock In \emph{Proceedings of the Conference of the North American Chapter
  of the Association for Computational Linguistics: Human Language Technologies
  (NAACL)}, pages 1185--1195, Atlanta, Georgia. Association for Computational
  Linguistics.

\bibitem[{Faruqui et~al.(2016)Faruqui, Tsvetkov, Neubig, and
  Dyer}]{faruqui:2016:infl}
Manaal Faruqui, Yulia Tsvetkov, Graham Neubig, and Chris Dyer. 2016.
\newblock \href {http://www.aclweb.org/anthology/N16-1077} {Morphological
  inflection generation using character sequence to sequence learning}.
\newblock In \emph{Proceedings of the Conference of the North American Chapter
  of the Association for Computational Linguistics: Human Language Technologies
  (NAACL)}, pages 634--643, San Diego, California. Association for
  Computational Linguistics.

\bibitem[{Heigold et~al.(2017)Heigold, Neumann, and van
  Genabith}]{heigold-neumann-vangenabith:2017:EACLlong}
Georg Heigold, Guenter Neumann, and Josef van Genabith. 2017.
\newblock \href {http://www.aclweb.org/anthology/E17-1048} {An extensive
  empirical evaluation of character-based morphological tagging for 14
  languages}.
\newblock In \emph{Proceedings of the Conference of the European Chapter of the
  Association for Computational Linguistics (EACL)}, pages 505--513, Valencia,
  Spain. Association for Computational Linguistics.

\bibitem[{Hulden et~al.(2014)Hulden, Forsberg, and
  Ahlberg}]{hulden-forsberg-ahlberg:2014:EACL}
Mans Hulden, Markus Forsberg, and Malin Ahlberg. 2014.
\newblock Semi-supervised learning of morphological paradigms and lexicons.
\newblock In \emph{Proceedings of the Conference of the European Chapter of the
  Association for Computational Linguistics}, pages 569--578, Gothenburg,
  Sweden. Association for Computational Linguistics.

\bibitem[{Kirov et~al.(2018)Kirov, Cotterell, Sylak-Glassman, Walther,
  Vylomova, Xia, Faruqui, Mielke, McCarthy, K{\"u}bler, Yarowsky, Eisner, and
  Hulden}]{KIROV18}
Christo Kirov, Ryan Cotterell, John Sylak-Glassman, G{\'e}raldine Walther,
  Ekaterina Vylomova, Patrick Xia, Manaal Faruqui, Sabrina~J. Mielke, Arya
  McCarthy, Sandra K{\"u}bler, David Yarowsky, Jason Eisner, and Mans Hulden.
  2018.
\newblock Unimorph 2.0: {U}niversal morphology.
\newblock In \emph{Proceedings of the Ninth International Conference on
  Language Resources and Evaluation (LREC)}. European Language Resources
  Association (ELRA).

\bibitem[{Nicolai et~al.(2015)Nicolai, Cherry, and
  Kondrak}]{nicolai-cherry-kondrak:2015:NAACL-HLT}
Garrett Nicolai, Colin Cherry, and Grzegorz Kondrak. 2015.
\newblock \href {http://www.aclweb.org/anthology/N15-1093} {Inflection
  generation as discriminative string transduction}.
\newblock In \emph{Proceedings of the Conference of the North American Chapter
  of the Association for Computational Linguistics: Human Language Technologies
  (NAACL)}, pages 922--931, Denver, Colorado. Association for Computational
  Linguistics.

\bibitem[{Nivre et~al.(2016)Nivre, de~Marneffe, Ginter, Goldberg, Hajic,
  Manning, McDonald, Petrov, Pyysalo, Silveira, Tsarfaty, and
  Zeman}]{NIVRE16.348}
Joakim Nivre, Marie-Catherine de~Marneffe, Filip Ginter, Yoav Goldberg, Jan
  Hajic, Christopher~D. Manning, Ryan McDonald, Slav Petrov, Sampo Pyysalo,
  Natalia Silveira, Reut Tsarfaty, and Daniel Zeman. 2016.
\newblock Universal dependencies v1: A multilingual treebank collection.
\newblock In \emph{Proceedings of the Tenth International Conference on
  Language Resources and Evaluation (LREC)}, Paris, France. European Language
  Resources Association (ELRA).

\bibitem[{Petrov et~al.(2012)Petrov, Das, and McDonald}]{PETROV12.274}
Slav Petrov, Dipanjan Das, and Ryan McDonald. 2012.
\newblock A universal part-of-speech tagset.
\newblock In \emph{Proceedings of the Eight International Conference on
  Language Resources and Evaluation (LREC)}, Istanbul, Turkey. European
  Language Resources Association (ELRA).

\bibitem[{Spencer(1991)}]{spencer1991morphological}
Andrew Spencer. 1991.
\newblock \emph{Morphological Theory: {A}n Introduction to Word Structure in
  Generative Grammar}.
\newblock Wiley-Blackwell.

\bibitem[{Straka et~al.(2016)Straka, Haji\v{c}, and
  Strakov{\'a}}]{straka2016udpipe}
Milan Straka, Jan Haji\v{c}, and Jana Strakov{\'a}. 2016.
\newblock {UDPipe}: {T}rainable pipeline for processing {CoNLL-U} files
  performing tokenization, morphological analysis, {POS} tagging and parsing.
\newblock In \emph{LREC}.

\bibitem[{Sylak-Glassman(2016)}]{sylak2016composition}
John Sylak-Glassman. 2016.
\newblock The composition and use of the universal morphological feature schema
  ({U}nimorph schema).
\newblock Technical report, Johns Hopkins University.

\end{thebibliography}
\bibliographystyle{acl_natbib}

\end{document}